\documentclass[a4paper]{article}

\usepackage[english]{babel}
\usepackage[utf8x]{inputenc}
\usepackage[T1]{fontenc}

\usepackage{algorithm}
\usepackage{algorithmic}
\usepackage{enumitem} 
\usepackage{amsmath}
\usepackage{amssymb}
\usepackage{graphicx}
\usepackage[colorinlistoftodos]{todonotes}
\usepackage[colorlinks=true, allcolors=blue]{hyperref}
\usepackage{color}
\usepackage{multirow}
\usepackage[round, sort,comma,authoryear]{natbib}
\usepackage{siunitx} 
\usepackage{verbatim}
\usepackage{tikz}
\def\checkmark{\tikz\fill[scale=0.4](0,.35) -- (.25,0) -- (1,.7) -- (.25,.15) -- cycle;}

\newcommand{\footnoteremember}[2]{
\footnote{#2}
  \newcounter{#1}
  \setcounter{#1}{\value{footnote}}
}
\newcommand{\footnoterecall}[1]{
\footnotemark[\value{#1}]
}

\title{A language processing algorithm for predicting tactical solutions to an operational planning problem under uncertainty}

\author{Eric Larsen\footnoteremember{cirrelt}{Department of Computer Science and Operations Research and CIRRELT, Universit\'e de Montr\'eal} \and Emma Frejinger\footnoterecall{cirrelt}\footnote{Corresponding author. Email: emma.frejinger@cirrelt.ca}}

\date{\today}

\begin{document}
\maketitle

\begin{abstract}
This paper is devoted to the prediction of solutions to a stochastic discrete optimization problem. Through an application, we illustrate how we can use a state-of-the-art neural machine translation (NMT) algorithm to predict the solutions by defining appropriate vocabularies, syntaxes and constraints. We attend to applications where the predictions need to be computed in very short computing time -- in the order of milliseconds or less. The results show that with minimal adaptations to the model architecture and hyperparameter tuning, the NMT algorithm can produce accurate solutions within the computing time budget. While these predictions are slightly less accurate than approximate stochastic programming solutions (sample average approximation), they can be computed faster and with less variability.
\end{abstract}

Keywords: integer linear programming, load planning, operational planning, tactical planning, supervised learning, neural machine translation

\section{Introduction}

Our research project is situated at the nexus of Operations research (OR) and Machine learning (ML). OR methods can be applied to a wide range of hard, discrete optimization decision problems and are crucial in numerous contexts throughout the world, e.g., for transportation planning, production management or control of energy systems. Fast solution algorithms are generally available for the deterministic problems whose characteristics are known exactly. However, most actual problems are naturally stochastic and much harder and time-consuming to solve, whence large-scale applications are typically restricted to deterministic formulations despite a lack of realism. 

Our objective is to demonstrate the application and performance of a state-of-the-art neural machine translation (NMT) algorithm for generating fast and accurate predictions of solutions to a stochastic discrete optimization decision problem. We build upon the methodology delineated in \cite{LarsEtAl18} and assume (i) that we can compute solutions to a decision problem under full information using an existing deterministic optimization model and a solver and (ii) that information about the problem is revealed  progressively, full information being available only at the time when the decision problem is ultimately solved. We wish to predict certain characteristics of this fully informed solution, based on currently available \emph{partial} information, calling such a characterization a \emph{tactical solution description}. Faced with this \emph{stochastic optimal prediction} problem, we predict the tactical solution descriptions using supervised ML, where the training data consists of a large number of fully informed problems that have been solved independently and off line. Crucially, we attend to applications where the predictions must be delivered with high accuracy and speed, actually much faster than solving a single decision problem under full information.

Reaching beyond the roots shared with \cite{LarsEtAl18}, we focus on tasks where the tactical solution and the available information are more detailed. 
Whereas \cite{LarsEtAl18} concentrates on the class of feedforward neural networks mapping between vectors of fixed length, we extend the scope to the class of neural machine translation (NMT) approximators. The latter generically map between ordered sequences of variable length and, through specialization, also map between any two types of data representations among the following: fixed-length vector, bag of fixed or variable size, (ordered) sequence of fixed or variable length. For comparative purposes, we also present a faster, albeit less precise baseline model built upon a data transformation and a standard feedforward neural architecture. 

Our closest neighbor in the OR-ML literature is \cite{VinyEtAl15} that applies supervised training to an NMT predictor in order to approximate solutions to a particular class of \emph{deterministic} integer linear programming (ILP) problems. \cite{NairEtAl18} also addresses the prediction of ILP solutions under imperfect information. However, its methodology, based on reinforcement learning, requires simulations and cannot deliver predictions at a speed that is sufficient for the applications that we wish to address. For a survey of the nexus of OR and ML, see \cite{BengEtAl18}.

Our motivating application concerns booking decisions of intermodal containers on double-stack trains: The assignment of containers to slots on railcars is a combinatorial optimization problem – called the load planning problem (LPP) – that cannot be solved deterministically at the time of booking due to imperfect information: the LPP depends on characteristics of both railcars (e.g., weight capacity, geometric loading restrictions), and containers (e.g., weight and size), and container weights are not available at the time of booking. Whereas \cite{LarsEtAl18} predicted how many containers of each type in a given set can be loaded on a given set of railcars of different types and how many of the railcars of each type are needed, here we predict with greater detail the number of containers of each type to be loaded on each railcar. In view of the intended real-time application in a high volume booking system, the solution must be computed in very short time (fraction of a second). This problem is of practical importance in the particular railway management regime of \emph{blocking with reclassification} and features characteristics that make it useful for illustrating the proposed methodology: Although an ILP formulation of the problem can be solved under full information by commercial ILP solvers in seconds to minutes \citep{MantEtAl17}, this formulation cannot be used directly for the application since the container weights are not known at the time of booking. Furthermore, for the purposes of booking decisions, the operational solution (assignment of each container to positions on railcars) is unnecessarily detailed.

\paragraph{\textbf{Contributions}}
The paper offers these contributions to the OR-ML literature:

\begin{itemize}
\item Demonstrate the use of a state-of-the-art NMT model for predicting solution descriptions of generic combinatorial decision problems under imperfect information in very short computation time.
\item Illustrate how to represent the combinatorial problem statement and solution description with the input and output vocabularies and syntaxes of an NMT model.
\item Indicate how to enforce constraints restricting either the input-output map or the output itself throughout training, validation and testing of the NMT model, as well as at prediction time, with a probability mask.
\item Define a measure of the discrepancy between predicted and actual solution descriptions that is invariant to order.
\item Present a baseline model built upon a data transformation and a standard feedforward neural architecture.
\item Illustrate the proposed methodology and perform a detailed analysis of its predictive and computational performance through an application to the container-railcar LPP.
\end{itemize}

The remainder of the paper is structured as follows:
Section 2 presents the related literature. It defines the NMT approximators and delineates their role in approximating the prediction function relating the statement of an optimal prediction problem and its solution. Section 3 adapts the NMT approximator to the container-railcar LPP problem. Chiefly, this involves specifying input and output languages and syntaxes, imposing constraints relating inputs and outputs and defining an appropriate measure of the predictive accuracy. Section 4 examines the computational properties of the NMT approximators in their application to the LPP problem. First, Sections 4.1 to 4.4 lay the required groundwork by describing successively the ML apparatus used in training and validating the NMT approximators, the container-railcar LPP data, the baseline NMT approximator and a lower bound for the predictive error based on an approximate stochastic programming solution. Finally, Section 4.5 discusses extensively the predictive performances and computation times achieved by alternative versions and implementations of the NMT and baseline approximators.

\section{Related Literature} \label{sec:literature}

We define formally the optimal prediction problem and explain the role of NMT approximators in approximating the prediction function relating the statements of the optimal prediction problem and their solutions. We also present the NMT model lying at the core of the main NMT approximator under examination.

\subsection{The Optimal Prediction Problem} \label{sec:pbDescription}
We briefly state the optimal prediction problem of \cite{LarsEtAl18} and highlight the scope of our work. Let a particular instance of a fully informed (deterministic) optimization problem be represented by the input feature vector ${\textbf{x}}$. The optimal fully detailed (i.e., that containing values of all decision variables), fully informed solution  is ${\mathbf{y}}^*({\mathbf{x}})
 :\equiv \arg \inf_{{\mathbf{y}} \in \mathcal{Y}({\mathbf{x}})} C({\mathbf{x}}, {{\mathbf{y}}})$, where $C({\mathbf{x}}, {\mathbf{y}})$ and $\mathcal{Y}({\mathbf{x}})$ denote respectively the cost function and the admissible space. We define a partition ${\textbf{x}} = [{\textbf{x}}_{\text{a}}, {\textbf{x}}_{\text{u}}]$ where ${\textbf{x}}_{\text{a}}$ contains available features and ${\textbf{x}}_{\text{u}}$ unavailable ones at the time of prediction. Furthermore, we denote by $g(\cdot)$ the mapping from the fully detailed and informed solution to the tactical solution description 
 featuring the level of detail relevant to the context at hand. Hence, $g({\mathbf{y}})$ is the synthesis of the fully detailed and informed solution ${\mathbf{y}}$ according to the tactical solution description embedded in $g(\cdot)$. Our goal is to approximate the solution $\bar{\mathbf{y}}^*({\mathbf{x}}_{\text{a}})$ to the following two-stage, \emph{optimal prediction stochastic programming} \citep[see, e.g.,][]{KallWall94, BirgLouv11, ShapEtAl09} problem:
 
 \begin{equation}
\bar{\mathbf{y}}^*({\mathbf{x}}_{\text{a}}) : \equiv \arg \inf_ {\bar{\mathbf{y}}({\mathbf{x}}_{\text{a}}) \in \bar{\mathcal{Y}}({\mathbf{x}}_{\text{a}})} \Phi_{{\mathbf{x}}_{\text{u}}} \lbrace \lVert
\bar{\mathbf{y}}({\mathbf{x}}_{\text{a}}) - g({\mathbf{y}}^*({\mathbf{x}}_{\text{a}}, {\mathbf{x}}_{\text{u}})) \rVert \mid {\mathbf{x}}_{\text{a}} \rbrace
\label{prediction first stage}
\end{equation}

\begin{equation}
{\mathbf{y}}^*({\mathbf{x}}_{\text{a}}, {\mathbf{x}}_{\text{u}}) :\equiv \arg \inf_{{\mathbf{y}} \in \mathcal{Y}({\mathbf{x}}_{\text{a}}, {\mathbf{x}}_{\text{u}})} C({\mathbf{x}}_{\text{a}}, {\mathbf{x}}_{\text{u}}, {\mathbf{y}})
\label{prediction second stage}
\end{equation}
where $\lVert \rVert$ denotes a suitable norm (e.g. the $L_1$- or $L_2$-norm when the output has fixed size) and $\Phi_{{\mathbf{x}}_{\text{u}}} \lbrace \lVert .\rVert \mid {\mathbf{x}}_{\text{a}} \rbrace$ denotes either the expectation or a quantile (e.g., the median) operation over the distribution of ${\mathbf{x}}_{\text{u}}$, conditional upon ${\mathbf{x}}_{\text{a}}$. Hence, $\bar{\mathbf{y}}^*({\mathbf{x}}_{\text{a}})$ is the optimal prediction of the synthesis of the second-stage optimizer $g({\mathbf{y}}^*({\mathbf{x}}_{\text{a}}, {\mathbf{x}}_{\text{u}}))$, conditionally on information available at first stage. Finally, $\mathcal{Y}({\mathbf{x}}_{\text{a}}, {\mathbf{x}}_{\text{u}})$ is the admissible space defined by the set of constraints relevant to the fully informed context, whereas $\bar{\mathcal{Y}}({\mathbf{x}}_{\text{a}})$ is defined only by the set of constraints relevant to the partially informed context.

We aim to generate a prediction function that can take any value of ${\mathbf{x}}_{\text{a}}$ as input, and outputs accurate predictions $\widehat{\mathbf{y}}^*({\mathbf{x}}_{\text{a}})$ of $\bar{\mathbf{y}}^*({\mathbf{x}}_{\text{a}})$. The difficulty of this task largely depends on the complexity of $g(\cdot)$. In comparison with \cite{LarsEtAl18}, we consider the challenging case of a more detailed tactical solution description. Here, the predictions are given by $\widehat{\mathbf{y}}^*({\mathbf{x}}_{\text{a}}) \equiv f({\textbf{x}}_{\text{a}};\boldsymbol{\theta})$ where $f(\cdot;\cdot)$ is a particular NMT-based approximator and $\boldsymbol{\theta}$ is a vector of tuned parameters.

We select $f(\cdot;\cdot)$ and $\boldsymbol{\theta}$ through supervised ML based on  input-output data made up of $({\textbf{x}}, \bar{\mathbf{y}}^*({\mathbf{x}}))$ pairs. Practically, treating the stochasticity in ${\textbf{x}}_{\text{u}}$ hinges on the particular definition of the data pairs that are used for learning. Similar to \cite{LarsEtAl18}, we use an implicit method by 
simply passing the dataset $\{({\mathbf{x}}^{(i)}_{\text{a}}, g({\mathbf{y}}^*({\mathbf{x}}^{(i)}_{\text{a}}, {\mathbf{x}}^{(i)}_{\text{u}})),~i=1,\ldots,m\}$ to ML. \cite{LarsEtAl18} already generated this data through a controlled probabilistic sampling of ILP problem statements (input) and by computing their solutions (output) with a standard ILP solver.

\subsection{NMT Approximators}
\label{nmt approximators}
Although its original purpose was to perform NMT between natural languages, the NMT apparatus can generically map between finite input and output sequences of either fixed or variable length, either ordered or unordered, whose elements take values in finite sets. As far as we know, \cite{SutsEtAl14} was the first to characterize the NMT apparatus as a generic approximator, introducing the expression ``sequence to sequence learning''. Hence, we find the term ``NMT approximator'' appropriate.

\subsubsection{NMT Models as Building Blocks of NMT Approximators}
\label{nmt models}

The chain rule of probabilities is the foundation of the NMT approximator. It expands the joint probability of the complete output sequence, conditionally upon the input sequence, as the product of the marginal probabilities of the single output elements, conditionally upon the previous output elements and the complete input sequence. NMT seeks a solution to $\arg \max_{\mathbf{y}} P(\mathbf{y}|\mathbf{x})$  where 
\begin{equation}
 P(\mathbf{y}|\mathbf{x}) \equiv \prod_{i=2}^IP(\mathbf{y_i}| \mathbf{y}_{i-1},..., \mathbf{y}_1; \mathbf{x}) P(\mathbf{y}_1| \mathbf{x})
\label{expanded_probs}
\end{equation}
is the probability of output sequence $\mathbf{y}$ in the target language conditionally upon input sequence $\mathbf{x}$ in the source language and $\mathbf{y}_i$ denotes element $i$ of the output sequence. An NMT model is simply a ML representation for $P(\mathbf{y}_i| \mathbf{y}_{i-1},..., \mathbf{y}_1; \mathbf{x})$. For a survey of NMT models, see \cite{YounEtAl17}.

Using the notation introduced in Section~\ref{sec:pbDescription}, we can write \eqref{expanded_probs} for our problem as
\begin{equation}
 P(\mathbf{y}| {\mathbf{x}}_{\text{a}}, \mathbf{\theta}) \equiv \prod_{i=2}^IP(\mathbf{y}_i| \mathbf{y}_{i-1},..., \mathbf{y}_1; {\mathbf{x}}_{\text{a}}, \mathbf{\theta})  P(\mathbf{y}_1| {\mathbf{x}}_{\text{a}}, \mathbf{\theta})
\label{expanded_probs_spec}
\end{equation}

\begin{equation}
\widehat{\mathbf{y}}^*({\mathbf{x}}_{\text{a}})  = \arg\max_{\mathbf{y}} P(\mathbf{y}| \mathbf{x_a}, \mathbf{\theta}).
\end{equation}
Standard supervised training, validation and testing can be used to select a particular class of NMT models, the associated hyperparameters and the parameter values $\mathbf{\theta}$. Recall that in our case the data consists of pairs of input and output sequences $\{({\mathbf{x}}^{(i)}_{\text{a}},
g({\mathbf{y}}^*({\mathbf{x}}^{(i)}_{\text{a}}, {\mathbf{x}}^{(i)}_{\text{u}})),~i=1,\ldots,m\}$. 

To generate the prediction of an output sequence given an input sequence, a search procedure successively selects individual elements of the output so as to maximize the joint probability of occurrence of the predicted output sequence conditionally on the input sequence. Calculations are based on \eqref{expanded_probs_spec}. The required marginal conditional probabilities are calculated with the trained NMT model $P(\mathbf{y}_i| \mathbf{y}_{i-1},..., \mathbf{y}_1; {\mathbf{x}}_{\text{a}}, \mathbf{\theta})$. Beam search (breadth-first search with limited memory) is widely used in the search procedures in view of its excellent performance and modest computational requirements.

Finally, we note that probability masks attached at each step in the search process to $P(\mathbf{y}_i| \mathbf{y}_{i-1},..., \mathbf{y}_1; {\mathbf{x}}_{\text{a}}, \mathbf{\theta})$ can be used to enforce constraints restricting either the map between $\mathbf{x}$  and $\mathbf{y}$ or only $\mathbf{y}$. General constraints can be enforced in this manner. We devote Section~\ref{constraints} to a discussion on how we impose constrains in our application.

\subsubsection{A Specific NMT Model}
The particular NMT model that we implement in our application is essentially the recurrent neural network (RNN) model of \cite{BahdEtAl14}, in which we introduce a number of additions and adaptations pertaining to the choice of the input and output vocabularies and syntaxes, as well as to the imposition of input-output constraints restricting the output syntax. Its architecture comprises the following elements:
\begin{enumerate}
\item Encoder: A neural network mapping the sequence of inputs into a sequence of latent ``annotations'' of same length. Annotations are meant to summarize the content of the input sequence and are made up of the concatenation of forward-stepping and backward-stepping states. Forward and backward encoding states are each generated with an RNN, equipped with gated recurrent units (GRU) whose role is to control how much the previous encoding state should be updated or propagated given the current input element.
\item Attention mechanism: A neural network responsible for producing a custom weighted average (a.k.a. ``context'') of all encoding annotations at each step in the generation of the output sequence. Its weights (a.k.a. ``alignments'') depend on the previous decoding state (defined below) and all encoding annotations. The attention mechanism is meant to flexibly exploit all annotations originating from the encoder for generating the current output element. Whereas the original model comprises a single attention mechanism, we also experimented with an extended architecture comprising multiple specialized encoders and attention mechanisms.
\item Decoder: A neural network mapping the sequence of current contexts produced by the attention mechanism and the sequence of previous output elements into a sequence of decoding states. The decoding states are generated with an RNN equipped with GRUs controlling how much the previous decoding state should be updated or propagated given the current context and the previous output element.
\item Output layer: A neural network mapping the current context and the past decoding state into the probability distribution of the current output element conditional upon input sequence and past output elements.
\end{enumerate}

The introduction of an attention mechanism -- the seminal contribution of \cite{BahdEtAl14} -- was responsible for bringing NMT models to the forefront of automatic translation. This innovation resulted in the first end-to-end NMT model to attain state-of-the-art performance in automatic translation (previously achieved with sequential statistical translation models) from English to French and English to German. It has spawned an abundant literature whose contributions range from straightforward alterations to major reinterpretations of the attention mechanism and inclusion into radically different neural architectures that exclude RNNs. See, e.g., the literature on so-called transformers initiated by \cite{VaswEtAl17}.

Overall, the improvements in performance achieved so far in the field of natural language translation by the successors to \cite{BahdEtAl14} have been modest. However, replacing the RNNs can lead to substantial reductions in computational times by allowing parallelization. RNNs lack in this regard. The sizes of the sets where the input and output sequences take their values (i.e. the vocabularies) are considerably smaller in our application than they are in natural language processing (less than 200 vs several thousands). Hence, the lack of parallelization resulting from the use of RNNs presents a smaller disadvantage in our application. Transformer NMT model may be considered for real-time calculations requiring even more speed than that can be afforded by the model in \cite{BahdEtAl14} or those among its followers whose architectures are based on RNNs.

\section{Adapting the NMT approximator to the Load Planning Problem}

We adapt the NMT approximator based on \cite{BahdEtAl14} to the container-railcar LPP. Whereas \cite{LarsEtAl18} predicts how many railcars of each type are used and how many containers of each type are loaded, we aim to predict \emph{loadings} of \emph{individual} railcars. A loading is a tuple specifying the type of a railcar that is being used and the numbers of containers of each type that are assigned to it. 
Predictions featuring this level of detail cannot be accomplished with the simpler predictive models considered in \cite{LarsEtAl18}.

\begin{figure}
    \centering
    \includegraphics[width=\textwidth]{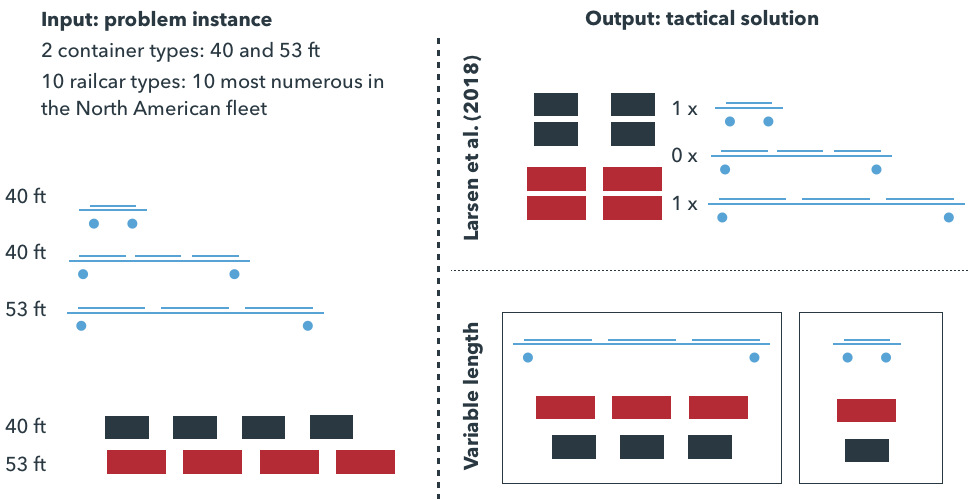}
    \caption{A small example instance illustrating input-output structure}
    \label{fig:input-output_example}
\end{figure}

We provide an illustrative example in Figure~\ref{fig:input-output_example}. Inputs of the NMT approximator are identical to those of \cite{LarsEtAl18}. Namely, they state the numbers of available railcars of each type and the numbers of assignable containers of each type. This example features three types of railcars and two types of containers. The particular instance shown includes one available railcar of each type and four assignable containers of each type. Our application covers the ten most common railcar types in North America, accounting together for close to 90\% of the rail car fleet, and two types of containers. The right-hand side of Figure~\ref{fig:input-output_example} illustrates the output. At the top, the output of \cite{LarsEtAl18} states the numbers of assigned containers of each type and the numbers of used railcars of each type. At the bottom, the output of the NMT approximator is of variable length and indicates for each used railcar how many of each type of container are to be loaded on it (without specifying the actual assignment to places on the railcar which would correspond to the operational solution). 

The output of the NMT approximator is a sequence of variable length with elements taking values in a finite set (since there are finitely many railcar types and there are for each type finitely many ways to load it with containers). For our purposes, the output sequence does not need to be ordered. However, ordering of input and output sequences is integral to the architecture of an NMT model. Hence, we simply disregard the order of the loadings in the predictions that are generated with the model and tally the count of each type of loading in the generated output sequence. 


The application of the NMT-based conditional predictor to the container-railcar LPP requires that LPP statement and solution description be expressed in the input and output languages. Essentially, we need to select appropriate vocabularies and syntaxes for the input and the output. 
Ceteris paribus, small vocabularies and simple, regular syntaxes are desirable in order to exploit the available training data efficiently. Any outstanding constraint restricting the map between LPP statement and solution or the LPP solution itself must also be upheld by the output syntax of the NMT. We discuss the details in the following two sections.

\subsection{Vocabularies and Syntaxes}

The NMT input phrase describing an LPP statement is made up of 26 tokens. The first 20 tokens are divided into 10 pairs. Each pair is associated with a particular type of railcar indexed from type 0 to type 9. Using a decimal representation, a pair specifies with two digits the number of available railcars of a particular type. Hence, the number of available railcars can range from 0 to 99 based on this syntax. In a similar fashion, the last 6 words in the input phrase are divided into 2 triples. Each triple is associated with a particular length of container, that is either 40~ft or 53~ft. Using a decimal representation, a triple specifies with three digits the number of assignable containers of this length. Hence, the number of assignable containers can range from 0 to 999. A distinct set of tokens representing digits from 0 to 9 is assigned to each railcar type and to each container length. Hence, the input vocabulary is made up of $10 * 10 + 2 * 10 = 120$ tokens.

Notice how a syntax based on the digits of a numerical representation implicitly modulates the effective length of the input sequence in relation with the number of represented objects: only higher input counts will have variable, non zero digits in the higher powers of their representation. This contributes to maintain sufficient statistical leverage throughout the input and output range \citep[see][p.~3]{VinyEtAl16}.


The output phrase describing the LPP solution description is based on a vocabulary of size 157. Notwithstanding empty loads, the examination of the admissible loading patterns for the 10 railcar types considered in our application indicates that there are in total 155 ways in which some 40~ft and 53~ft containers can be loaded onto the 10 types of railcars. Hence, there are 155 tokens associated with these 155 distinct feasible loadings. There are two additional tokens in the output vocabulary. The token ``EOS'' terminates every output phrase. The token "BLANK" signifies that no container was loaded onto any railcar. The latter cannot be preceded by any other token and must be followed by EOS. Hence, an output sequence must match either one of these cases:
\begin{itemize}
\item Variable number of output tokens belonging to the 155 tokens describing the feasible loadings, terminated by the token EOS.
\item BLANK followed by EOS. 
\end{itemize}

\subsection{Constraints Restricting the Output Syntax}
\label{constraints}

The marginal conditional probability distribution $P(\mathbf{y}_i| \mathbf{y}_{i-1},..., \mathbf{y}_1; {\mathbf{x}}_{\text{a}}, \mathbf{\theta})$  produced by the NMT model 
must assign zero probabilities to the following occurrences: The current token
\begin{itemize}
\item implies the assignment of a number of containers that exceeds the number of remaining assignable containers;

\item implies the use of an unavailable railcar,

\item BLANK follows another token,

\item BLANK is not followed by EOS,

\item EOS appears in first position,

\item EOS is not last in the output sequence.

\end{itemize}

These requirements may be enforced by updating and applying a probability mask in the output layer of the NMT model. The mask projects the probability of any excluded event to zero and remaining non-zero probabilities are normalized so as to sum to one. 
The application of constraints with a probability mask is also an alternative to the approach exemplified by \cite{VinyEtAl15} where the relevant constraints are embodied in the model (pointer network) architecture.

\subsection{Measuring the Predictive Accuracy}
The following statistic $D$ measures the mean \emph{unordered} discrepancy over the observations $i=1,\dots,I$ of a data set in the container-railcar LPP application between the actual output sequences and the corresponding predicted output sequences generated with a trained approximator.
\begin{equation}
D = \frac{\sum_{i=1}^I \sum_{j=1}^J \min_{\pi} \sum_{k=1}^{\widetilde{K}_{ij}}\sum_{l=1}^L | \widetilde{n}_{ij\pi(k)l}^{act} - \widetilde{n}_{ijkl}^{pred}|}{\sum_{i=1}^I \sum_{j=1}^J \sum_{k=1}^{\widetilde{K}_{ij}} \sum_{l=1}^L  \widetilde{n}_{ijkl}^{act}},
\end{equation}
where
\begin{itemize}
\item the index $j$  of a railcar type runs from $1$ to $J$ ($J=10$ in our application), 
\item the index $k$ of the \emph{padded}, actual and predicted loadings in observation $i$ for railcar type $j$ runs from $1$ to $\widetilde{K}_{ij}$,
$\widetilde{K}_{ij} = \max (K_{ij}^{act}, K_{ij}^{pred})$, 
\item $K_{ij}^{act}$ is the number of actual loadings in observation $i$ related to railcar type $j$ \emph{before padding}, 
\item $K_{ij}^{pred}$ is the number of predicted loadings in observation i related to railcar type $j$ \emph{before padding}, 
\item the index $l$ of a container length runs from $1$ to $L$ ($L=2$ in our application), 
\item $\pi$ acts as a permutation of the loading indexes $k=1,...,\widetilde{K}_{ij}$ for a given observation $i$ and a given railcar type $j$, 
\item $n_{ijkl}^{act}$ is the actual number of loaded containers of length $l$ for loading $k$ among the loadings related to railcar type $j$ in observation $i$ \emph{before padding}, 
\item $\widetilde{n}_{ijkl}^{act}$ is the \emph{padded}, actual number of loaded containers of length $l$ for loading $k$ amond the loadings related to railcar type $j$ in observation $i$; $\widetilde{n}_{ijkl}^{act} = n_{ijkl}^{act}, k = 1, ..., K_{ij}^{act}$ and $\widetilde{n}_{ijkl}^{act} = 0, k = (K_{ij}^{act} + 1), ...,\widetilde{K}_{ij}$ are \emph{padding}, empty loadings, 
\item $n_{ijkl}^{pred}$ is the predicted number of loaded containers of length $l$ for loading $k$ among the loadings related to railcar type $j$ in observation $i$ \emph{before padding}, 
\item $\widetilde{n}_{ijkl}^{pred}$ is the \emph{padded}, predicted number of loaded containers of length $l$ for loading $k$ amond the loadings related to railcar type $j$ in observation $i$; $\widetilde{n}_{ijkl}^{pred} = n_{ijkl}^{act}, k = 1, ..., K_{ij}^{pred}$ and $\widetilde{n}_{ijkl}^{pred} = 0, k =(K_{ij}^{pred} + 1), ...,\widetilde{K}_{ij}$ are \emph{padding}, empty loadings.
\end{itemize}

Practically, the inner expression $\min_{\pi} \sum_{k=1}^{\widetilde{K}_{ij}}\sum_{l=1}^L | \widetilde{n}_{ij\pi(k)l}^{act} - \widetilde{n}_{ijkl}^{pred}|$ can be translated into the following program that we solve with a general purpose integer programming solver, $\forall (i, j)$:

\begin{equation}
\min_a \sum_{u=1}^{\widetilde{K}_{ij}} \sum_{v=1}^{\widetilde{K}_{ij}} a_{uv} \sum_{l=1}^L | \widetilde{n}_{ijul}^{act} - \widetilde{n}_{ijvl}^{pred}|
\end{equation}

subject to:
\begin{equation}
\sum_u a_{uv} = 1, \forall v
\end{equation}

\begin{equation}
\sum_v a_{uv} = 1, \forall u
\end{equation}

\begin{equation}
a_{uv} \in \lbrace 0, 1 \rbrace, \forall (u, v)
\end{equation}

$D$  is invariant to permutations in the output sequence. This is as it should be since we are not concerned with the order of an output sequence for our purposes, but rather with the counts of the types of loadings appearing in the solution. $D$ could  not be used in the training process since it is not amenable to analytical gradient-based optimization. Although it could in principle have been used in the validation process, it was not, since its computation is too highly demanding for these purposes.

\section{Computational Results}

This section examines the computational properties of the NMT approximators in their application to the LPP problem. Sections \ref{ML_apparatus} to \ref{SAA} establish the required foundations. They successively describe the ML apparatus used in training and validating the NMT models, the container-railcar LPP data, the baseline NMT approximator and a lower bound for the predictive error based on solutions originating from approximate stochastic programming. Section \ref{predictive performance} discusses extensively the predictive performances and computation times achieved by alternative versions and implementations of the NMT and baseline approximators.

\subsection{The ML apparatus}
\label{ML_apparatus}

We trained the NMT and baseline models with pseudo-likelihood maximization and stochastic mini-batch gradient descent. Except when considering extension to multiple encoders and attention mechanisms, the sizes and numbers of layers in the NMT model were kept as in \cite{BahdEtAl14}. We experimented with the adadelta and adam methods \citep[see, e.g.,][]{Rude19} of automatic learning rate (i.e., step size) adaptation. Regularization was exerted through dropout and early stopping. The latter was applied with a patience of at least one epoch. Mini-batch size was equal to 64 throughout. The validation process applied in the selection of hyperparameter settings and for the purpose of early stopping was based on pseudo-likelihood. As explained in Section~\ref{nmt approximators}, predictions were generated with beam search. 

Training, validation and prediction generation operations were performed on a single high-power GPU (Nvidia V-100 or Titan XP). GPU RAM requirements for training, validation and prediction generation were  and in most cases well below 4 GB. We used the Python 2.7 programming language with the Theano 1.02 symbolic computation library \citep{Thea16} and the Groundhog meta-library \citep{Grou15}. The code made publicly available by the authors of Groundhog and \cite{BahdEtAl14} was adapted and extended \citep[see][]{Grou15}. The Cython C language compiler \citep{BehnEtAl11} ensured the fast calculation of the probability masks. Measurement of the final predictive performance was performed with Java and CPLEX on an Intel i7 processor.

\subsection{The Container-Railcar LPP Data}
\label{data}
Our data is similar to that of \cite{LarsEtAl18}. It is 
partitioned into four classes, as reported in Table~\ref{tab:dataClasses}. This partitioning facilitates experiments where models are trained and validated on easier instances (A) and tested on either similar, harder (B, C) or hardest ones (D). There are two data sets of class A: A' contains 10~M instances and has been randomly divided into training (64\%), validation (16\%) and test (20\%) sets. A'' is independent from A' and contains 100~K instances. There are additional independent 100~K data sets for each one of the other classes. Those are labeled respectively B'', C'' and D'' and are used in conducting extraneous testing.

\begin{table}[h]
\begin{center}
\begin{tabular}{|p{1cm} p{3.5cm} cc |}
\hline
Class name & Description & \# of containers & \# of platforms \\ \hline
A & Easiest ILP instances & [1, 150] & [1, 50] \\ 
B & More containers than A (excess demand) & [151, 300] & [1, 50] \\
C & More platforms than A (excess supply) & [1, 150] & [51, 100] \\
D & Largest and hardest instances & [151, 300] & [51, 100] \\ 
\hline
\end{tabular}
\caption{Data classes}
\label{tab:dataClasses}
\end{center}
\end{table}

\subsection{Baseline Model}
\label{baseline model}

State-of-the-art NMT models are complex and require considerable investments in knowledge and computational resources for their training, selection and implementation. We are interested in comparing the predictive performance and computational requirements associated with an NMT model \citep{BahdEtAl14} with those of a simpler, standard neural network. Our idea is as follows: since the size of the output vocabulary is quite small (157), it is possible to transform the data so that it in effect constitutes a sample of the marginal probability of the next loading, conditionally upon the input sequence and all previously committed loadings. Such a transformation would be intractable with the output vocabulary sizes typically encountered in natural language processing. The purpose of the baseline model is to build a baseline approximator along the lines explained in Section~\ref{nmt models}. The latter will be evaluated in terms of its own merit and will also act as a reference in appraising the usefulness of the NMT approximator.

\paragraph{Transforming the original data set} Consider a particular example in the original data set. The latter consists, on the one hand, of an input sequence specifying the numbers of available railcars of each type and the numbers of assignable containers of each length and, on the other hand, of an output sequence of variable length, say $L$, made up of individual railcar loadings. This particular original example can be expanded into $L$ new examples in the transformed data set. Each one of the $L$ new examples features an input sequence of a fixed length equal to $10 + 2 + 157 =  179$. The elements in this sequence stand for the 10 numbers of available railcars of each type, followed by the 2 numbers of assignable containers of each length, followed by 157 numbers specifying the numbers of committed loadings of each type, defined according to the original output vocabulary. All $L$ new examples share common first 12 elements in their input sequence. These 12 elements reflect the input sequence of the original example. Output of a new example consists of one loading type index among the 157. The indexes appearing in the outputs of the $L$ new examples correspond to the loadings appearing in the original output sequence of length $L$. Hence, the last 157 elements of the $L$ new examples from $0$ to $L-1$ of the transformed data set can be generated successively as follows. 

\begin{enumerate}

\item New example with index $0$ has zeros in all $L$ positions from 12 to 178 of its input sequence and its output is equal to the index of the first loading in the original output sequence.

\item Input sequence of new example with index $i, i>0$, results from input sequence of new example $i-1$ by adding 1 to the number appearing in position $12 + j$, where $j$ is the the output of new example $i-1$. Output of example $i$ is equal to index of the ith loading in the original output sequence according to the original output vocabulary.

\end{enumerate}

\paragraph{Feedforward Architecture} 
Based on the transformed data set, a standard classification model can be trained so as to approximate the marginal distribution of the next loading, conditional upon the original input sequence and the loadings previously committed. For this purpose, we selected a feedforward neural network with 179 units in the input layer and 157 units in the softmax output layer. All units were equipped with rectified linear activations (ReLU). A small number of alternative configurations in a range of numbers of hidden layers and units per hidden layer were considered. The constraints described in Section~\ref{constraints} above were imposed during training, validation and generation. A probability mask similar to that of the NMT model was applied for this purpose.

\subsection{A Lower Bound for the Predictive Error}
\label{SAA}

The stochastic limits of the statistic $D$ that are estimated in Section~\ref{predictive performance} for the stated combinations of model, training algorithm, training set and validation or testing set are all bounded from below and away from zero due to the stochastic nature of the prediction problem. Estimates of the relevant lower bounds can be calculated from approximate solutions of the corresponding optimal prediction stochastic programming problems. For this purpose, we use the sample average approximation (SAA) method. Such lower bounds make it possible to assess how far the NMT approximators are from performing optimally and if an attempt at improving their performance by increasing the capacity of the NMT model and/or the size of training data is worthwhile. We stress that approximate stochastic programming solutions such as SAA must be calculated for every particular value of ${\textbf{x}}_{\text{a}}$, whereas the NMT and baseline approximators define prediction functions that are valid for every value of ${\textbf{x}}_{\text{a}}$.

The SAA statistics reported in Table \ref{tab:SAA} are based on two-stage samples whose first stage is given by the 100~K examples of A'' and whose second stage comprise samples of container weights (i.e., scenarios) with sizes equal to 5, 10, 25, 50, 99. Since the estimated mean value of $D$ monotonically decreases from 0.108 to 0.079 at a decreasing rate as a function of the sample size (i.e., number of scenarios) and is identical for sample sizes 50 and 99, it is reasonable to view 0.079 as being in the vicinity of the lower bound for the optimal prediction problem. We shall refer to this value as the \emph{SAA bound} over the set A''. Table \ref{tab:SAA} also reports estimates of the means and standard deviations of the times required to compute the SAA solution for each second stage sample size. We note that, as expected, the computing times and the associated standard errors increase with the number of scenarios while the standard error of $D$ decreases.

Calculation of an individual SAA solution to the LPP proceeds through these steps: for a given set of first stage variables (inputs whose values are available in advance in the application), draw a sample of sets of second stage variables (inputs whose values are unavailable in advance). Using an ILP solver, compute LPP solutions for each combination of the set of first stage values and one of the individual sets of second stage values. Minimize the sample average of $D$ over all LPP solutions thus computed. The resulting minimizer is the sought SAA solution.

\begin{table}[!htbp]
    \centering
    \begin{tabular}{|l|cc|cc|}
    \hline
    Nb. of scenarios$^{*}$ & \multicolumn{2}{c|}{$D$}  & \multicolumn{2}{c|}{Computing time [s]} \\ 
        & Est. mean & Est. std dev & Est. mean & Est. std dev \\\hline
    5 & 0.089 & 0.166 & 4.649 & 15.841 \\
      & (5.26E-03) & & (5.01E-02) & \\\hline
    10 & 0.084 & 0.157 & 9.250 & 18.187 \\
      & (4.96E-04) & & (5.75E-02) & \\\hline
    25 & 0.081 & 0.152 & 23.182 & 29.817 \\
      & (4.8E-04) & & (9.43E-02) & \\\hline
    50 & 0.079 & 0.150 & 46.539 & 54.130 \\
      & (4.73E-04) & & (1.71E-01) & \\\hline
    99 & 0.079 & 0.148 & 93.277 & 109.972 \\
      & (4.69E-04) & & (3.48E-01) & \\\hline
     \multicolumn{4}{l}{\small * number of sets of weights (scenarios) drawn for each example}\\
      \multicolumn{4}{l}{\small Standard error of estimate is reported between parentheses.}\\
      \end{tabular}
    \caption{Properties of the SAA predictor over set A''}
    \label{tab:SAA}
\end{table}


\subsection{Predictive Performance and Computing Times}
\label{predictive performance}
Table~\ref{tab:pred_erros} reports estimates of the mean and standard deviation of the predictive error $D$ achieved over a number of sets by the model trained and validated over the 6.4~M training examples and 1.6~M validating examples of data set A'. The first row reports the predictive performance over the 1.6~M validating examples of data set A'. Data set A'' in second row is independently drawn from the same distribution as A' and the predictive performance over this set can be compared to the SAA bound equal to 0.079. Distributions of data sets B'', C'' and D'' differ from that used for generating training and validation data and the predictive performance over these sets indicates the extent to which a model trained and validated over easiest examples found in data set A'' can generalize to the harder and hardest examples in data sets B'', C'' and D''. Table~\ref{tab:pred_times} reports estimates of the mean and standard deviation of the corresponding prediction times.

The results show that the prediction error incurred by the NMT approximator over data independent from but similar to the data used for training and validation is in the vicinity of the the SAA bound and therefore quite good. While there is still room for improvement, it must be noted that the NMT approximator reaches this performance with very little hyperparameter tuning beyond the final values reported in \cite{BahdEtAl14}. Furthermore, the estimated standard deviation of $D$ is small and, importantly, estimated mean prediction times are in the order of milliseconds, hence within our limited budget, and estimated standard deviations of prediction times are small. The estimate of the mean error incurred by the SAA predictor over the set A'' is smaller than that of the MNT approximator, even when using as few as 5 scenarios. However, the estimated mean computing time is several orders of magnitudes longer and with larger standard deviation. In comparison to the NMT model, the baseline model exhibits a surprisingly good performance and estimated mean computing times of the baseline approximator are even shorter than those of the NMT approximator. 

\begin{table}[!htbp]
    \centering
    \begin{tabular}{|l|cc|cc|}
    \hline
    Data & \multicolumn{2}{c|}{NMT 1-attention} & \multicolumn{2}{c|}{Baseline} \\ 
    \# examples & Est. mean $D$ & Est. std dev $D$ & Est. mean $D$ & Est. std dev $D$ \\\hline
    A' 1.6~M & 0.101 & 0.169 &  0.162 & 0.218 \\
    & (1.33E-04) & & (1.72E-04) &
    \\ \hline
    A'' 100~K & 0.115 & 0.179 &  0.172 & 0.222
    \\
    & (5.65E-04) & & (7.00E-04) &
    \\
    B'' 100~K & 0.054 & 0.127 &  0.063 & 0.144
    \\
    & (4.03E-04) & & (4.56E-04) &
    \\
    C'' 100~K & 0.257 & 0.197 &  0.270 & 0.168
    \\
    & (6.24E-04) & & (5.31E-04) &
    \\
    D'' 100~K & 0.181 & 0.178 &  0.096 & 0.156
    \\
    & (5.63E-04) & & (4.95E-04) &
    \\ \hline
     \multicolumn{4}{l}{\small Standard error of estimate is reported between parentheses.}\\
    \end{tabular}
    \caption{Prediction errors for constrained models trained on 6.4~M examples of A', beam search width of 5, adadelta method, single attention NMT model}
    \label{tab:pred_erros}
\end{table}

\begin{table}[!htbp]
    \centering
    \begin{tabular}{|l|cc|cc|}
    \hline
    Data & \multicolumn{2}{c|}{NMT 1-attention} & \multicolumn{2}{c|}{Baseline} \\ 
    \# examples & Est. mean & Est. std dev & Est. mean & Est. std dev    \\\hline
    A' 1.6~M & 0.025 & 1.55E-02 & 0.010 & 6.07E-03 \\ 
    & (1.23E-05) & & (4.80E-06) &  \\ \hline
    A'' 100~K & 0.025  & 1.52E-02 & 0.009 & 5.34E-03 \\
    & (4.82E-05) &  & (1.69E-05) & \\
    B'' 100~K & 0.031 & 1.71E-02 & 0.010 & 6.25E-03 \\
    & (5.42E-05) & & (1.98E-05) & \\
    C'' 100~K & 0.047 & 2.91E-02 & 0.016 & 8.88E-03 \\
    & (9.20E-05) & & (2.81E-05) & \\
    D'' 100~K & 0.091 & 3.02E-02 & 0.028 & 6.63E-03 \\
    & (9.55E-05) & & (2.10E-05) & \\
    \hline
    \multicolumn{4}{l}{\small Standard error of estimate is reported between parentheses.}\\
    \end{tabular}
    \caption{Prediction times in seconds}
    \label{tab:pred_times}
\end{table}

A number of findings emerged from the experiments leading to the selection of the best performing NMT approximator. For instance, we found in general that predictive performance was unaffected by increasing the beam width from 5 to 10 but that computing time was one order of magnitude larger on average.

While input-output constraints were always enforced at prediction time, we could choose to enforce them or not during training and validation. We achieved similar final predictive performance in both cases. However, it frequently caused numerical instability when truncating the probabilities at prediction time. As a result, we applied input-output constraints throughout training, validation and generation.

Table~\ref{tab:ctimes} reports the time required for training the NMT and baseline models over the 6.4~M examples of A' whether the constraints are imposed or not and whether the adadelta or adam step size adaptation methods are applied. Training with adam when constraints were enforced was slower and required on average more than twice as much time as adadelta to reach the threshold of early stopping than when constraints were not enforced. Overall, the first four rows of the table illustrate that the benefits of the step size adaptation methods are context-specific and that they must viewed as any other hyperparameter.

\begin{table}[!htbp]
    \centering
    \begin{footnotesize}
    \begin{tabular}{|llcrrr|}
    \hline
         Model & Step & Constraints & time [s]  & \# epochs & Total \\
         & & (training) & / mini-batch & & time [h] \\ \hline
         NMT 1-attention & adadelta & \checkmark & 0.5 & 10.3 & 143 \\
         NMT 1-attention & adadelta & & 0.36 & 15.32 & 153 \\ \hline 
         NMT 1-attention & adam & \checkmark & 0.832 & 16.32 & 377 \\
         NMT 1-attention & adam & & 0.68 & 6 & 113 \\ \hline 
         NMT 3-attention & adadelta & \checkmark & 0.90 & 8 & 200 \\
         NMT 3-attention & adadelta & & 0.70 & 11 & 214 \\ \hline
         Baseline & adadelta & \checkmark & 0.016 & 6$^*$ & 2.5 \\
         \hline
         \multicolumn{6}{l}{* of transformed data, approximately 0.6 epochs of original data}
    \end{tabular}
    \end{footnotesize}
    \caption{Training times over 6.4~M examples A'}
    \label{tab:ctimes}
\end{table}


The predictive performances reported in Table~\ref{tab:pred_erros} are based on models trained on 6.4~M examples of class A'. In order to assess the impact of sample size on predictive performance, we trained the NMT and baseline models on 640~K and 64~K examples of class A'. On the one hand, a ten-fold reduction in the size of the training set had no detrimental effect on predictive performance (measured over A', estimate of mean value of $D$ equals 0.101 with an estimated standard error of 1.29E-04). On the other hand, total training time until early stopping was considerably smaller, decreasing from 143~h to 50.5~h. However, a further reduction in the size of the training set to 64~K examples was consequential as the estimate of mean value of $D$ measured over A' increased to 0.146 (with estimated standard error of estimate equal to 1.68E-04). 

We found that the single attention NMT model instance resulting from training and validation over respectively 6.4~M and 1.6~M examples of A' achieves on the training set a pseudo-likelihood value nearly identical to the pseudo-likelihood value achieved over the validation set (-0.000602 vs -0.000604, respectively). This points strongly to the conclusion that improvements in predictive performance, if any are possible, should be sought by expanding model capacity rather than the size of the training set. The latter is corroborated by the close value of -0.000644 measured over the same validation set when the training data set was tentatively doubled in size to 12.8~M.

As an attempt to usefully increase capacity, we implemented a multiple attention NMT model featuring three encoder-attention pairs, instead of the single one found in \cite{BahdEtAl14}. One pair focuses on the input elements related to the available railcars, another focuses on the input elements related to the assignable containers and a third one attends to the whole input sequence as in \cite{BahdEtAl14}. This failed to improve upon the predictive performance achieved by the original single-attention model (on class A, estimate of mean value of $D$ equals 0.101 for the multiple attention model with estimated standard error of estimate equal to 1.30E-04). Moreover, as verified in Table~\ref{tab:ctimes}, training times for the three-attention model are considerably longer than the single-attention model.

As a quality control, we were interested in measuring the performance achieved by the NMT approximator when predicting simpler solution descriptions as in \cite{LarsEtAl18}. The performance criterion in this case is the sum of the mean absolute prediction error over the number of used slots and number of loaded containers. The lower bound associated with this statistic could be computed with stochastic programming (SAA). It is equal to 0.82 (with an estimated standard error equal to 0.0087) whereas, depending on the size of the training set, the values reported for the statistic in \cite{LarsEtAl18} for a classification model are respectively 0.965 (standard error of 0.002) and 1.481 (standard error of 0.018). We find it reassuring that the NMT approximator achieves for this statistic a value of 1.421 (standard error of 0.003) over a similar test set with very little hyperparameter tuning.

\section{Conclusion}
This paper has addressed the problem of computing a close approximate solution to any particular instance of a stochastic discrete optimization problem in a fraction of a second. The difficulty of this problem depends on the desired level of detail of the solution. Building upon \cite{LarsEtAl18}, we focused on predicting more detailed solution descriptions requiring machine learning models that can handle outputs of variable size. We aimed to verify whether a state-of-the-art neural machine translation (NMT) algorithm could generate fast and accurate predictions of solutions with minimal adaptation to the model architecture and minimal hyperparameter tuning. We adapted the existing NMT models by introducing problem specific input and output vocabularies and syntaxes and input-output constraints. While these are specific to our application, we believe that similar vocabularies, syntaxes and input-output constraints can be tailored to a broad range of programming problems, for instance to bin packing problems. The proposed NMT and baseline approximators demonstrated good predictive performances that are in the neighborhood of the stochastic lower bound computed with sample average approximation. Their comparative advantage with respect to sample average approximation is that of very fast computation with very low standard deviation.

\section*{Acknowledgements}
We are grateful to Yoshua Bengio who was involved in an initial discussion about this problem. We have also benefited from discussions with Andrea Lodi. 
This research was funded by the Canadian National Railway Company (CN) Chair in Optimization of Railway Operations at Université de Montréal and a Collaborative Research and Development Grant from the Natural Sciences and Engineering Research Council of Canada (CRD-477938-14). The research is also partially funded by the ``IVADO Fundamental Research Project Grants'' under project entitled ``Machine Learning for (Discrete) Optimization''. Computations were made on the supercomputers Briarée and Guillimin, managed by Calcul Qu\'ebec and Compute Canada. The operation of these supercomputers is funded by the Canada Foundation for Innovation (CFI), the Minist\`ere de l'\'Economie, de la Science et de l'Innovation du Qu\'ebec (MESI) and the Fonds de recherche du Qu\'ebec - Nature et technologies (FRQ-NT). We also benefited from the computing resources provided by Mila.

\bibliographystyle{plainnat_custom}
\bibliography{References}

\end{document}